%% file: paper.tex
\documentclass[runningheads,a4paper]{llncs}

\usepackage{amssymb}
\usepackage{amsmath}
\usepackage{graphicx}
\usepackage[table,pdftex]{xcolor}
\usepackage{xspace}
\usepackage[colorlinks=true,citecolor=blue,urlcolor=black]{hyperref}
\usepackage{lineno}
\usepackage{subcaption}
\usepackage[font={small}]{caption}

\usepackage{booktabs}

\definecolor{colornotes}{rgb}{0.6,0,0}

\makeatletter
\newcommand{\specificthanks}[1]{\@fnsymbol{#1}}% Inserts a specific \thanks symbol
\makeatother

\begin{document}

\mainmatter  % start of an individual contribution
\title{Unsupervised domain adaptation in brain lesion segmentation with adversarial networks}
\titlerunning{Unsupervised domain adaptation with adversarial networks}

\author{
Konstantinos Kamnitsas\inst{1,4}\thanks{Part of this work was carried on when KK was an intern at Microsoft Research.} \and
Christian Baumgartner\inst{1} \and
Christian Ledig\inst{1} \and \\
Virginia Newcombe\inst{2,3} \and
Joanna Simpson\inst{2} \and
Andrew Kane\inst{2} \and
David Menon\inst{2,3} \and
Aditya Nori\inst{4} \and
Antonio Criminisi\inst{4} \and
Daniel Rueckert\inst{1} \and
Ben Glocker\inst{1}
}
\authorrunning{Kamnitsas et al.}
\institute{
Biomedical Image Analysis Group, Imperial College London, UK \and
Division of Anaesthesia, Department of Medicine, Cambridge University, UK \and
Wolfson Brain Imaging Centre, Cambridge University, UK \and
Microsoft Research Cambridge, UK
}

\maketitle

%\linenumbers %remove before submission

\vspace{-5mm}
\input{sections/abstract}
\input{sections/introduction}

\vspace{-2mm}
\input{sections/method}
\input{sections/experiments}
\input{sections/conclusion}
\vspace{-1mm}
\input{sections/acknowledgements}

\bibliographystyle{splncs03}
\bibliography{cites}

\end{document}

%% file: sections/abstract.tex
% !TeX root=../paper.tex

\begin{abstract} Significant advances have been made towards building accurate automatic segmentation systems for a variety of biomedical applications using machine learning. However, the performance of these systems often degrades when they are applied on new data that differ from the training data, for example, due to variations in imaging protocols. Manually annotating new data for each test domain is not a feasible solution. In this work we investigate unsupervised domain adaptation using adversarial neural networks to train a segmentation method which is more invariant to differences in the input data, and which does not require any annotations on the test domain. Specifically, we learn domain-invariant features by learning to counter an adversarial network, which attempts to classify the domain of the input data by observing the activations of the segmentation network. Furthermore, we propose a multi-connected domain discriminator for improved adversarial training. Our system is evaluated using two MR databases of subjects with traumatic brain injuries, acquired using different scanners and imaging protocols. Using our unsupervised approach, we obtain segmentation accuracies which are close to the upper bound of supervised domain adaptation.

\let\thefootnote\relax\footnote{Email correspondence to: konstantinos.kamnitsas12@imperial.ac.uk}

\end{abstract}

%% file: sections/introduction.tex
% !TeX root=../paper.tex

\section{Introduction}

Great advancements have been achieved in machine learning, particularly with supervised learning algorithms, reaching human-level performance on applications that a few years ago would be considered extremely challenging. However, a common assumption in machine learning is that training and test data are drawn from the same probability distribution \cite{Valiant1984}. Methods are trained on data from a \textit{source domain} $D_S=\left\{\mathcal{X}_S,P(X_S)\right\}$, where $\mathcal{X}_S$ is a feature space, $X_S=\left\{x_{S1},...,x_{Sn}\right\}, x_{Si} \in \mathcal{X}_S$ the data and $P(X_S)$ the marginal distribution that their features follow. In an image segmentation problem, for example, $X_S$ could be samples (voxels or patches) from multi-spectral MR scans, $\mathcal{X}_S$ is the feature space defined by the available MR sequences and $P(X_S)$ is the distribution of intensities in the sequences. In the developing stage of a supervised algorithm, given corresponding ground truth labels $Y_S=\left\{y_{S1},...,y_{Sn}\right\}, y_{Si} \in \mathcal{Y}_S$, such as segmentation masks, where $\mathcal{Y}_S$ the label space, a predictive function $f_S(x) = P_S(y|x)$ is learnt via training and configuration of hyper-parameters on the data ($X_S,Y_S$). $f_S(\cdot)$ tries to approximate the optimal function $f^{\prime}_{S}(x), x \in \mathcal{X}_S$ that generated $Y_S$. At the time of deployment, however, these methods often under-perform or fail if the testing data come from a different \textit{target domain} $D_T=\left\{\mathcal{X}_T,P(X_T)\right\}$, with $\mathcal{X}_T \neq \mathcal{X}_S$ and/or $P(X_T) \neq P(X_S)$. This is because the optimal predictive function $f^{\prime}_{T}(x), x \in \mathcal{X}_T$ for $D_T$ may differ from $f^{\prime}_{S}(\cdot)$, and so the learnt $f_S(\cdot)$ will not perform well on $D_T$. The above scenario is common in biomedical applications due to variations in image acquisition, in particular, in multi-center studies. Training and testing data may differ in contrast, resolution, noise levels ($P(X_T) \neq P(X_S)$) or even type of sequences ($\mathcal{X}_T \neq \mathcal{X}_S$). Despite the rapid advancements in representation learning, this issue has been shown to affect even the latest models \cite{ullman2016atoms}. Generating labelled databases is time consuming and often expensive, and assuming annotations for training are available for each new domain is neither realistic nor scalable. Instead, it is desired to develop methods that can learn from existing databases and generalize well or adapt to the target domain without the need for additional training data.

Transfer learning (TL) \cite{pan2010survey} investigates development of predictive models by leveraging knowledge from potentially different but related domains and tasks. Even between tasks where label spaces $\mathcal{Y}_S$ and $\mathcal{Y}_T$ differ, TL can take advantage of similarities in the underlying structure of the mappings $f_S:\mathcal{X}_S \mapsto \mathcal{Y}_S$ and $f_T:\mathcal{X}_T \mapsto \mathcal{Y}_T$. A subclass of TL is \textit{multi-task learning}, where a model is trained on multiple related tasks \textit{simultaneously}. Most related to this work, \textit{domain adaptation} (DA) is the subclass of TL that assumes $\mathcal{Y}_S = \mathcal{Y}_T$ and only the domains differ. It explores learning a function $f_a(\cdot)$ that performs well on both domains, under the basic assumption that such a function exists \cite{ben2010theory}.

%We assume $Y_S$ were generated by an unknown function $f^{\prime}_{S}(x)$, $x \in X_S$.
In this work we investigate \textit{unsupervised domain adaptation} (UDA) \cite{jiang2008literature}. In this setting we assume the availability of a labeled database $S=(X_S,Y_S)$ from source domain $D_S$, along with an \textit{unlabeled} database $T=(X_T)$ from a different but related target domain $D_T$. We wish to model the unknown optimal function $f^{\prime}_{T}(\cdot)$ for labelling $X_T$. However since no labels are available for $D_T$, $f^{\prime}_{T}(\cdot)$ cannot be learnt. This is in contrast to supervised DA, which requires at least some labelled data for $D_T$. Instead, we try to learn a representation $h_a(x)$ that maps $X_S$ and $X_T$ to a feature space that is invariant to differences between the two domains, as well as a function $f_{ah}(\cdot)$ learnt using data $\left\{ X_S,Y_S,X_T \right\}$, such that $f_a(x) = f_{ah}(h_a(x))$ approximates $f^{\prime}_{S}(\cdot)$ and is closer to $f^{\prime}_T(\cdot)$ than any function $f_S(\cdot)$ that can be learnt using only the source data $(X_S,Y_S)$. 

% Instead, we try to learn a representation $h_a(x)$ that maps $X_S$ and $X_T$ to a feature space such that a classifier $f_a(x) = f_{ah}(h_a(x))$ learnt using data $\left\{ X_S,Y_S,X_T \right\}$ is closer to $f_T(x)$ than any function $f_S(\cdot)$ that can be learnt using only the source data $(X_S,Y_S)$. 

\textbf{Contributions:} In this work we develop a domain adaptation method based on adversarial neural networks \cite{ganin2016domain,goodfellow2014generative}. We propose the adversarial training of a segmenter and a domain-classifier, which aims to make the representation learnt by the segmenter invariant to domain-specific factors. We describe and analyse the development of domain-adversarial networks for the purpose of segmentation, which to the best of our knowledge has not been previously performed. We investigate the adaptation of layers at various depths and propose multi-connected adversarial networks, which we show improve domain adaptation. We employ our system for the segmentation of traumatic brain injuries (TBI), investigating adaptation between databases acquired using two different scanners with difference in the available MR sequences. We show that without utilizing any labels in the target domain, our method closes the performance gap with respect to supervised learning with target labels to a large extent.

\textbf{Related Work:} TL and DA have attracted significant interest over the years. Comprehensive reviews of early works can be found in \cite{pan2010survey,ben2010theory,jiang2008literature}. Popularity of TL increased with the wide adoption of neural networks when their features were found to be effective when transferred across tasks. For example, features learnt from natural images were used off-the-shelf for detecting peri-fissural nodules \cite{ciompi2015automatic}. More commonly, TL is performed via pre-training on a source task, followed by fine-tuning for the target task via supervised training \cite{shin2016deep}. A representative example of TL via multi-task learning was presented in \cite{moeskops2016deep}. A network was trained simultaneously for segmentation of brain tissue, pectoral muscle and coronary arteries. These experiments show that much of a network's capacity can be shared between a variety of tasks. Note, all of the above require labels in $D_T$.

In contrast, DA explores the case where label spaces $(Y_S,Y_T)$ are the same and little or no labelled data is available in $D_T$. In \cite{van2015transfer} the authors explored supervised DA with SVM-based adaptive classifiers in the scenario where source and target data are acquired with different protocols. This method, however, requires labelled target data. Unsupervised DA was tackled in \cite{heimann2013learning} via instance weighting, but this relies on strong assumptions about the data distributions. \cite{bermudez2016scalable} performed UDA with boosted decision stumps with a search for visual correspondences between source and target samples. This is not as flexible as our approach nor scales well to large databases. The authors in \cite{bermudez2016scalable} question the feasibility of DA with neural networks on 3D data due to memory requirements. Here, we show that using adversarial 3D networks is indeed a viable approach.

%% file: sections/method.tex
% !TeX root=../paper.tex

\section{Unsupervised domain adaptation with adversarial nets}
\label{sec:method}

The accuracy of a binary classifier that distinguishes between samples from two domains can serve as a proxy of the divergence of distributions $P(X_S)$ and $P(X_T)$, which otherwise is not straightforward to compute. This idea was first introduced in \cite{ben2010theory}. Inspired by this, the authors of \cite{ganin2016domain} presented a method for simultaneously learning a domain-invariant representation and a task-related classifier by a single neural network. This is done by minimizing the accuracy of an auxiliary network, a domain-discriminator, that processes a hidden representation of the main network and tries to classify the domain of the input sample. This approach formed the basis of our work. We below describe its extension for segmentation and our proposed multi-connected system.

\input{figures/method/figSystem}

\subsection{Segmentation system with domain discriminator}

\textbf{Segmenter:} At the core of our system is a fully convolutional neural network (CNN) for image segmentation \cite{Long2014}. Given an input $x$ of arbitrary size, which can be a whole image or a sub-segment, this type of network predicts labels for multiple voxels in $x$, one for each stride of the network's receptive field over the input. The parameters of the network $\mathbf{\theta}_{seg}$ are learnt by iteratively minimizing a segmentation loss $\mathcal{L}_{seg}$ using stochastic gradient descent (SGD). The loss is commonly the cross-entropy of the predictions on a training batch $B_{seg}=\left\{(x_{1},y_{1}),...,(x_{N_{seg}},y_{N_{seg}})\right\}$ of $N_{seg}$ samples. In our settings, $(x_i,y_i)$ are sampled from the source database $S = (X_S,Y_S)$, for which labels $Y_S$ are available. We borrowed the 3D multi-scale CNN architecture from \cite{kamnitsas2016efficient}, depicted in Fig~\ref{fig:system} and adopt the same configuration for all meta-parameters.

\textbf{Domain discriminator:} When processing an input $x$, the activations of any feature map (FM) in the segmenter encode a hidden representation $h(x)$. If samples come from different distributions $P(X_S) \neq P(X_T)$, e.g. due to different domains, and the filters of the segmenter are not invariant to the domain-specific variations, the distributions of the corresponding activations will differ as well, $P(h(X_S)) \neq P(h(X_T))$. This is expected when the segmenter is trained only on samples from $S$ where learnt features will be specific to the source domain. Similar to \cite{ganin2016domain}, we choose a certain representation $h_a(x)$ from the segmenter and use a second network as a domain-classifier that takes $h_a(x)$ as input and tries to classify whether it comes from $P(h_a(X_S))$ or $P(h_a(X_T))$. This is equivalent to classifying the domain of $x$. Classification accuracy serves as an indication of how source-specific the representation $h_a(\cdot)$ is. The architecture we use for a domain classifier is a 3D CNN with five layers. The first four have 100 kernels of size $3^3$. The last classification layer uses $1^3$ kernels. This architecture has a receptive field of $9^3$ with respect to its input $h_a(\cdot)$ and was chosen for compatibility with the size of feature maps in the 3 last layers of the segmenter.

We train this domain-discriminator simultaneously with the segmenter. For this, we form a second training batch $B_{adv} = \left\{ (x_{1},y_{1}^d),...,(x_{N_{adv}},y_{N_{adv}}^d)\right\}$. Equal number of samples $x_i$ are extracted from $X_S$ and $X_T$, so there is no bias towards either. $y_{i}^d$ is a label that encodes the domain of $x_i$, used as the training target. $B_{adv}$ is processed by the segmenter, at the same time with $B_{seg}$ or interleaved to lower memory requirements, computing activations $h_{a}(x) \forall x \in B_{adv}$. These activations are then processed by the discriminator, which classifies the domain of each sample in $B_{adv}$. The discriminator's classification loss $\mathcal{L}_{adv}$ is minimized through optimization of the parameters $\theta_{adv}$. 

A complication arises for the joint training. The samples from $S$ are shared in an SGD iteration for the two losses in the algorithm of \cite{ganin2016domain}. However, many segmentation methods use weighted sampling in order to mitigate class-imbalance, for example by oversampling rare classes \cite{kamnitsas2016efficient,havaei2016brain,moeskops2016deep}. Such sampling requires segmentation masks that are not available for $T$ whose samples are extracted randomly. In this case, the discriminator should not compare those against non-randomly extracted samples from $S$, as it could easily associate activations for the over-weighted classes with domain $S$ and fail to learn useful domain-discriminative features. Hence, we resort to forming entirely separate batches. $B_{adv}$ is formed of 20 image segments, randomly extracted from images in $S$ and $T$. As done in \cite{kamnitsas2016efficient}, weighted sampling is used for extracting 10 segments from $S$ to form $B_{seg}$. This ensures countering of class-imbalance for the segmenter, while being unbiased on the samples used for the discriminator.

\textbf{Domain adaptation via adversarial training:} We aim at adapting the representation $h_a(\cdot)$ to become invariant to variations between $S$ and $T$. To this end, we expose the accuracy of the domain-discriminator to the segmenter and let it alter its parameters such that its FMs that comprise $h_{a}(\cdot)$ do not contain cues about the input domain. This is done by incorporating the domain-discriminator's loss $\mathcal{L}_{adv}$ into the training objective of the segmenter, which now aims to simultaneously maximize the domain classification loss and minimize the segmentation loss $\mathcal{L}_{seg}$, or:
\begin{equation} \label{eq:segAdaptLoss}
\mathcal{L}_{segAdv}(\theta_{seg}) = \mathcal{L}_{seg}(\theta_{seg}) - \alpha \mathcal{L}_{adv}(\theta_{seg})
\end{equation}
$\alpha$ is a positive weight that defines the relative importance of the domain-adaptation task for the segmenter. This optimization is possible with regular SGD, as the adversarial networks are interconnected and gradients of $\mathcal{L}_{adv}$ can propagate back through the discriminator and into the segmenter. This process was implemented in \cite{ganin2016domain} via a custom \textit{gradient-reversal layer}, which is not needed if the optimization is formulated as in Eq.~(\ref{eq:segAdaptLoss}), as also noted by the authors.

\subsection{Multi-connected adversarial networks}
\label{subsec:multiconn}

\iffalse
Some more intuition in the optimization process can be given if we separate the weights of the segmenter $\mathbf{\theta}_{seg} = \left\{ \mathbf{\theta}_{f}, \mathbf{\theta}_{c} \right\}$. $\mathbf{\theta}_{f}$ are the parameters up to the layer that the adversarial is connected and thus contribute to both terms in Eq.~\ref{eq:segAdaptLoss}. They can be considered as the feature extractors of the representation $h_a(\cdot)$. $\mathbf{\theta}_{c}$ are the weights of the segmenter's layers that are deeper than the feature maps where the adversarial is connected to, only affect the segmentation loss \ref{eq:segLoss} and can be considered as the classifier $f_{ah}(h_a(x))$. 

loss rewritten as: $\mathcal{L}_{segAdv}(\theta_{f},\theta_{c}) = \mathcal{L}_{seg}(\theta_{f},\theta_{c}) - \alpha \mathcal{L}_{adv}(\theta_{f})$
\fi

A natural question to arise concerns which layer(s) of the segmenter should be adapted. In \cite{tzeng2014deep}, the authors investigated which of the last three fully connected layers of an AlexNet leads to better accuracy when adapted via MMD \cite{borgwardt2006integrating}, concluding it is the last hidden layer that is optimal in their settings. Earlier layers are commonly not adapted as their features are considered rather generic and transferable across related tasks \cite{ganin2016domain,icml2015_long15}. 

We argue that adapting only the last layers might not be ideal, especially for the case of segmentation. The accuracy of classification networks depends mostly on high-level patterns. For precise segmentation, however, fine patterns such as detailed texture and small contrast variations are likely to be important. These fine patterns are extracted in early layers and are more susceptible to image-quality variations between domains. Adapting top layers makes them invariant to such variations, but its still a loss of capacity if such features have been already extracted by early layers, which may not be well adapted by the weakened adversarial gradients that reach them. On the other hand, if only early layers are adapted, assuming that the adaptation is not ideal and the features not entirely free of factors of variation between the two domains, the network could recover source-specific patterns at greater depth. For these reasons we propose an architecture where the domain discriminator is connected at multiple layers of the segmenter. First, this removes source-specific patterns early on but also disallows their recovery at deeper layers. Furthermore, the discriminator is enabled to process a large variety of features for discriminating between the domains, increasing its performance and thus the quality of the gradients for the domain adaptation. Finally, by seeing the whole adversarial network as an auxiliary cost function for the segmenter, this type of connections can be compared with deep-supervision \cite{lee2015deeply}, which allows better flow of the gradients incoming from $\mathcal{L}_{adv}$ throughout the segmenter and as such can improve learning of quality features. Our main results are based on feeding input $h_{in}(\cdot)$ to the discriminator from FMs of layers 4,6 and 8 of both high and low resolution pathways, as well as the 10-th hidden layer of the segmenter (cf. Fig.~\ref{fig:system}). After the FMs of the low resolution pathway are upsampled, all FMs are cropped to match the size of the deepest layer and concatenated. A detailed analysis of the effect of adapting different layers is presented in Sec.~\ref{subsec:analysis}.

%% file: figures/method/figSystem.tex
\begin{figure}[!h] 
\centering
\begin{subfigure}[b]{1.0\textwidth}
	\centering
	\includegraphics[clip=true, trim=0pt 0pt 0pt 0pt, width=1.0\textwidth]{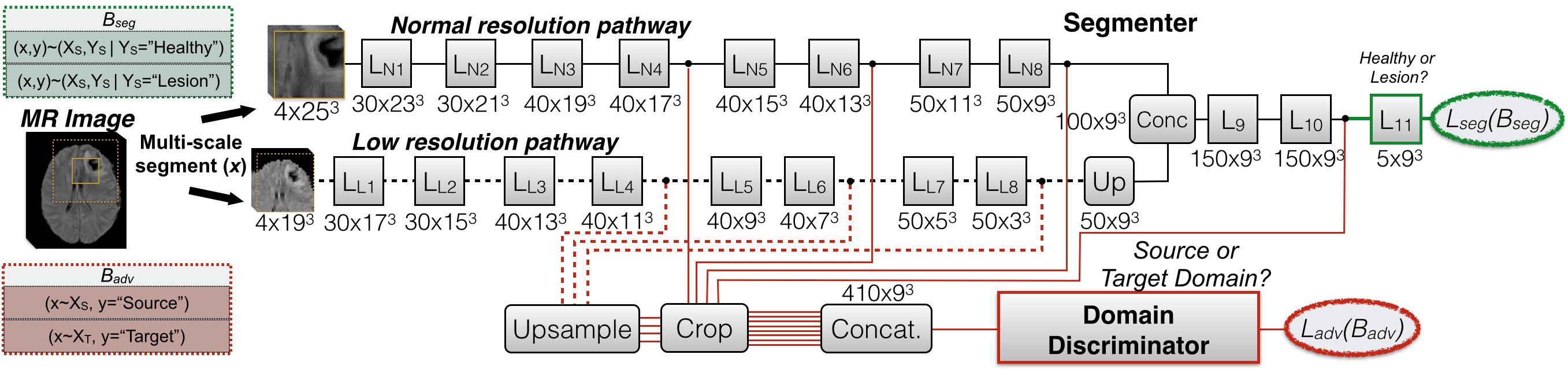}
	%\caption{}
	%\label{fig:connsB}
\end{subfigure}
\caption{Proposed multi-connected adversarial networks. Segmenter: we use the 3D CNN architecture presented in \cite{kamnitsas2016efficient}. Dashed lines denote low resolution features. Input samples are multi-modal, although not depicted. Discriminator: We use a second 3D CNN for classifying the domain of input $x$, by processing activations at multiple layers of the segmenter. Red lines show the path of the adversarial gradients, from $L_{adv}$ back to the segmenter. See text for details on architecture. }
\label{fig:system}
\vspace{-4mm}
\end{figure}

%% file: sections/experiments.tex
% !TeX root=../paper.tex

\section{Experiments}
\label{sec:experiments}
\vspace{-1mm}
\subsection{Material}
\vspace{-1mm}
We make use of two databases with multi-spectral MR brain scans of patients with moderate to severe TBI, acquired within the first week of injury. The first database consists of 61 subjects, imaged on a 3-T Siemens Magnetom TIM Trio. The MR sequences are isotropic MPRAGE (1mm$^3$), axial FLAIR, T2 and Proton Density (PD) (0.7$\times$0.7$\times$5mm), and Gradient-Echo (GE) (0.86$\times$0.86$\times$5mm). The second database consists of 41 subjects, imaged on a 3-T Siemens Magnetom Verio. This database includes MPRAGE, FLAIR, T2 and PD sequences, acquired at the same resolution as in the first database. The important difference is that instead of GE, a Susceptibility Weighted Image (SWI) is acquired (0.7$\times$0.7$\times$5mm). On both databases, all visible lesions were manually annotated on the FLAIR, GE and SWI by clinical experts. Here, we focus on binary segmentation of abnormalities within the brain tissue, and extra-cerebral pathologies were treated as background. All images are skull-stripped, resampled to isotropic 1mm$^3$ and affinely registered to MNI space. Image intensities under the brain masks are normalized to zero-mean and unit-variance, after windowing the lowest and top 2\% of the intensity histograms.
\vspace{-3mm}
\subsubsection{Source ($S$) and target ($T$) databases:} GE and SWI are commonly used in TBI studies due to their great sensitivity to haemorrhages, allowing detection of lesions not visible in other sequences (cf. Fig.~\ref{fig:qualitative}). SWI is a type of GE that offers greater sensitivity and image quality \cite{shenton2012review}. For the purpose of this study, the first database, with GE available, is considered the \textit{source} database $S$ used to train the segmenter in a supervised manner. The second database, with SWI available, is considered the \textit{target} database $T$ on which we aim to successfully apply the trained segmenter. This corresponds to a typical scenario where a training database is generated on data coming from one clinical site, and new test data coming from another site with varying protocol. Motivated by the similarity between GE and SWI, we will consider them as an interchangeable input channel to our segmentation system, unless stated otherwise. The difference between GE and SWI is contributing the largest variation between distributions $P(X_S)$ and $P(X_T)$, although some variation may come from differences between other sequences. Using our unsupervised domain adaptation, we aim to learn features that are invariant to these domain differences without the need for any annotations on the target domain. Treating different sequences as the same input is also considered in \cite{van2015transfer}, however, using a supervised adaptation approach.

\subsection{Configuration of the training schedule}
\label{subsec:configOfSystem}

A complication of adversarial training concerns the training schedule of the two connected networks, which influences the way they interact. The strength with which the segmenter is adapting its features in order to counter the domain-discriminator is controlled by the parameter $\alpha$ (cf. Eq.~(\ref{eq:segAdaptLoss})). We set $\alpha=0$ for the first $e_{1}=10$ epochs and let both networks learn independently. This allows the segmenter to initially learn features for the segmentation of $S$ without being influenced by noisy adversarial gradients from an initially poorly performing domain-discriminator. After epochs $e_{1}$, when the discriminator's performance has increased, we start countering it to learn domain invariant features with the segmenter. For this, we increase $\alpha$ according to the linear schedule $\alpha = \alpha_{max} \frac{e_{curr} - e_{1}}{e_{2} - e_{1}}$, where $e_{2}=35$ and $\alpha_{max}$ is the maximum weighting, so $\alpha$ equals $\alpha_{max}$ after epoch $e_{2}$. Finally, at epoch 43 we start refining the segmenter's features by gradually lowering its learning rate. The discriminator is optimized with constant learning rate 0.001. In the following, $\alpha_{max}=0.05$ is used. In Sec.~\ref{subsec:analysis} we present a sensitivity analysis showing robust behavior across a range of values for $\alpha_{max}$. $e_{1}$,$e_{2}$ and the total duration of this piecewise linear schedule were determined empirically for satisfactory convergence without prolonging training time. Optimal settings are not fully explored yet and may vary between different tasks and the relative difficulty of each network's specific task.

\subsection{Evaluation}
\label{subsec:eval}

We performed multiple experiments to obtain upper and lower bounds of baseline accuracy on the challenging task of TBI lesion segmentation. The experiments are discussed below, quantitative results are summarized in Tab.~\ref{tab:results} and examples of segmentations are given in Fig.~\ref{fig:qualitative}. For a fair comparison, the same 2-fold split of $T$ was used in all experiments that utilized annotated samples from $T$.

\begin{table}[!h]
\centering
\scriptsize
\caption{Comparison of our method's performance on $T$ with several baselines. Our system significantly closes the gap between the lower bound, when the segmenter is trained on $S$ only, and the upper bound, when the segmenter is also trained with labelled data from $T$. Values are given in format \textit{mean} (\textit{std}).}
\label{tab:results}
\begin{tabular}{@{}lllll@{}}
\toprule
\multicolumn{1}{c}{}				& DSC 	& Recall			& Precision			\\ \midrule
Train on S						& 15.7(13.5) \hspace{5mm} 	& 80.4(12.3) \hspace{5mm}		& 09.5(09.0)		\\
Train on S (No GE/SWI)			& 59.7(22.1)		& 55.7(22.6)		& 69.7(21.5)		\\
\midrule
\textbf{Train on S $\rightarrow$ UDA to T (ours)}	& 62.7(19.8)		& 58.9(21.2)		& 71.6(18.4)		\\
\midrule
Train on T						& 63.5(20.2)		& 60.6(21.1)		& 71.5(19.8)		\\
Train on S+T						& 66.5(17.7) 	& 66.6(19.1)		& 69.4(19.0)		\\
Train on S+T (GE/SWI diff chan.)\hspace{5mm}	& 64.7(19.2) 	& 65.7(20.2)		& 67.0(20.8)		\\
\bottomrule
\end{tabular}
\vspace{-3mm}
\end{table}

\textbf{Train on $S$, test on $T$:} We perform standard supervised training of the segmenter on $S$ without adaptation. To segment $T$, motivated by the similarity between GE and SWI sequences, at test time we use SWI in the channel used for GE during training. Even though these sequences can serve similar purposes in the analysis of TBI by radiologists, this approach totally fails, proving them not directly interchangeable as input to a CNN.

\textbf{Train on $S$ (No GE/SWI), test on $T$:} We repeat the previous experiment but only use the common sequences of $S$ and $T$ in both training and testing, neglecting GE and SWI. The experiment was repeated twice to reduce random variations between training sessions. This corresponds to a practical scenario, where we need to segment $T$ by only using annotated training data from $S$, and serves as the \textit{lower bound} of accuracy for our system.

\textbf{Train on $T$, test on $T$:} We perform a 2-fold validation using supervised training on half of $T$ and testing on the other half. We use all sequences of $T$. The obtained performance is similar to what was reported in \cite{kamnitsas2016efficient}, although on a different database. This experiment provides another indication for the expected accuracy on this challenging segmentation task. 

\textbf{Train on $S$ and $T$, test on $T$:} To obtain an \textit{upper bound} of accuracy, we train the segmenter on all data of $S$ and half the data of $T$, using their manual annotations. The same input channel is used for GE of $S$ and SWI of $T$. We then test on the other half of data from $T$. The experiment is repeated for the other split of $T$. We balance the samples from the two domains in each batch $B_{adv}$ to avoid biasing the segmenter towards $S$ that has more subjects. With supervised training on $T$, the system learns to interchange GE and SWI successfully. This setting uses all available data from both domains, both images and manual annotations, and serves as an estimate of optimal, supervised transfer learning.

\textbf{Train on $S$ and $T$, test on $T$ (GE/SWI in different channels):} We perform a sanity check that using GE and SWI in the same input channel is reasonable. We repeat the previous experiment but using a CNN with six channels, with separate ones for GE and SWI. The channel is filled with $-4$ when the sequence is not available, which corresponds to a very low value after our intensity normalization. From this the CNN learns when the sequence is missing and we found this to behave better than common zero-filling. The segmenter performs better than supervised training on $T$ only. This indicates that information from both domains is used. However, knowledge transfer is not as strong as when GE and SWI, which share much information, are used in the same channel.

\textbf{Proposed unsupervised domain adaptation:} We train the segmenter on all data of $S$ and adapt the domains using half the subjects of $T$, but no labels. GE and SWI share the same input channel. We test segmentation accuracy on the other half of $T$. The experiment is repeated for the other fold. Our method learns filters invariant to the two imaging protocols and transfers knowledge from $S$ to $T$, allowing the system to segment haemorrhages only visible on SWI without ever seeing a manual annotation from $T$ (Fig.~\ref{fig:qualitative}). This improves by 3\% DSC over the non-adapted segmenter that uses only information from $S$ and the common sequences, covering 44\% of the difference between the practical lower bound and the upper bound achieved by supervised domain adaptation using labels from both domains.

\input{figures/qualitative/figQualitative}
\vspace{-3mm}
\subsection{Analysis of system}
\label{subsec:analysis}

\subsubsection{Effect of adapting layers at different depths:}

\input{figures/0_figuresToUse/1_fig_connections/figConnectionsWithTable}

We investigate how the depth of the adapted layers affects our system. For this, we repeat the experiment with domain adaptation from $S$ to $T$, changing the layers from which input to the domain-discriminator is provided. Results are shown on Fig.~\ref{fig:connections} and Tab.~\ref{table:connections}. Note that we connect the discriminator at the same layers of both multi-scale pathways of the segmenter (for example, L4 means connections to the 4th layers of both pathways). Adaptation of early layers tends towards over-segmentation (increased recall but lower precision). It has been noticed that severe over-segmentation occurs without adaptation (Fig.~\ref{fig:qualitative}). These observations make us believe that the segmenter recovers source-specific features between the adapted and the classification layer. Comparing L2 and L(2,4,6,8,10) shows that this is alleviated by multiple connections that enforce domain invariance throughout the segmenter. Since, however, the behaviour of multi-connected adversarials is strongly defined by the shallowest connection, we avoid adapting the earliest layers which seems less beneficial but would slow down convergence.

\vspace{-3mm}
\subsubsection{Effect of adaptation's strength via $\alpha_{max}$:}

Here we investigate the sensitivity of our method with respect to $\alpha_{max}$, which defines how strongly the discriminator is countered by the segmenter. Fig.~\ref{fig:effectA} shows that higher values lead to quicker adaptation but the accuracy is rather stable for a significant range of values $\alpha_{max} \in [0.05,1.0]$. We note this range might differ for other applications and that smooth convergence is generally preferred for learning high quality features over steep schedules that alter the loss surface aggressively. Finally, we observe that strongly countering the discriminator does not guarantee better performance on $T$. A theoretical reason is that a more domain-invariant representation $h_a(x)$ likely encodes less information about $x$. This information loss increases the Bayes error rate and the entropy of the predictions by the learnt $f_a(x)=f_{ah}(h_a(x))$. After a certain level of invariance, this can outweigh the benefits of domain-adaptation \cite{ben2010theory,jiang2008literature}.

\input{figures/0_figuresToUse/2_fig_effectA/figEffectAInnerLegend}

%% file: figures/qualitative/figQualitative.tex
\begin{figure}[!h] 
\vspace{-2mm}
\centering
\begin{subfigure}[b]{1.0\textwidth}
	\centering
	\includegraphics[clip=true, trim=0pt 0pt 0pt 0pt, width=1.0\textwidth]{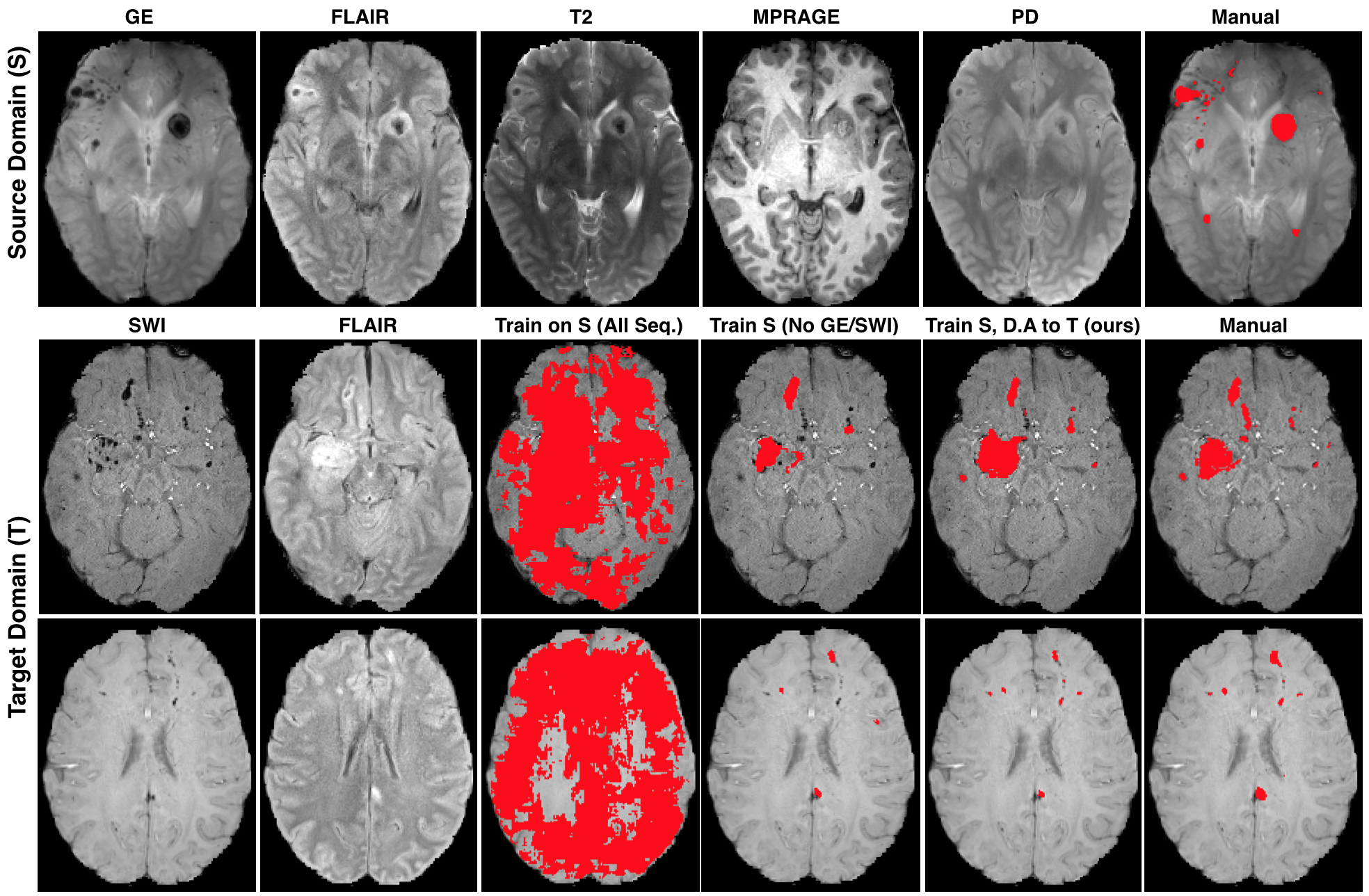}
	%\caption{}
	%\label{fig:connsB}
\end{subfigure}
\caption{(top row) Example case from S. (middle/bottom row) Visual results for two examples. A model trained on $S$ fails on $T$ when GE is simply replaced by SWI (3rd col.). A model trained on $S$ using only the four common sequences misses micro-bleeds visible only on SWI (4th col.). Our method mitigates these problems by learning features invariant to the imaging protocol (5th col.). (T2, MPRAGE and PD of $T$ are used but not depicted.)}
\label{fig:qualitative}
\vspace{-5mm}
\end{figure}

%% file: figures/0_figuresToUse/1_fig_connections/figConnectionsWithTable.tex
\begin{figure}[!h] 
\centering

\begin{subfigure}[b]{0.28\textwidth}
	\centering
	\includegraphics[clip=true, trim=0pt 0pt 0pt 0pt, width=1.0\textwidth]{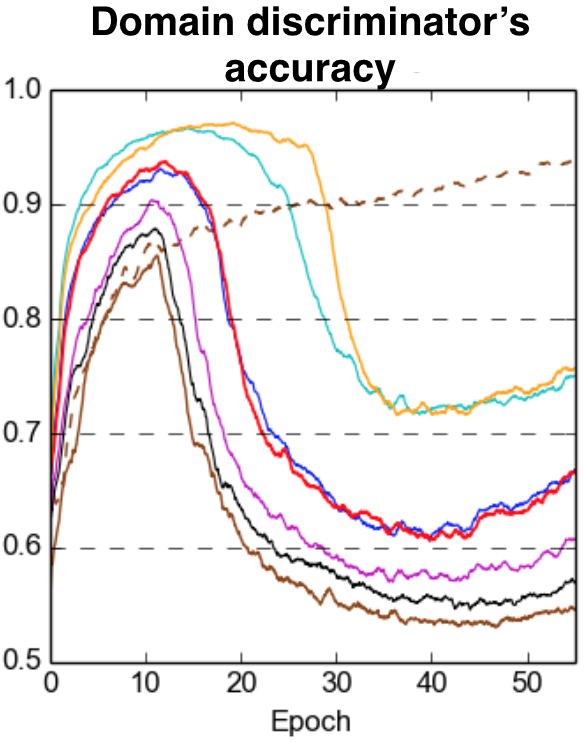}
	%\caption{}
	%\label{fig:connsA}
\end{subfigure}
\begin{subfigure}[b]{0.28\textwidth}
	\centering
	\includegraphics[clip=true, trim=0pt 0pt 0pt 0pt, width=1.0\textwidth]{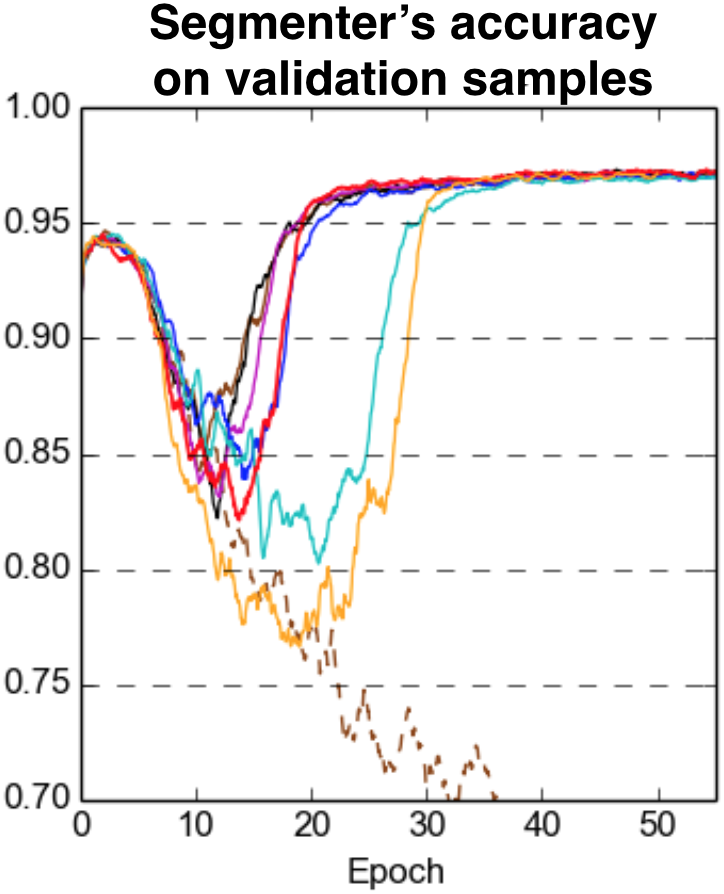}
	%\caption{}
	%\label{fig:connsB}
\end{subfigure}
\begin{subfigure}[b]{0.28\textwidth}
	\centering
	\includegraphics[clip=true, trim=0pt 0pt 0pt 0pt, width=1.0\textwidth]{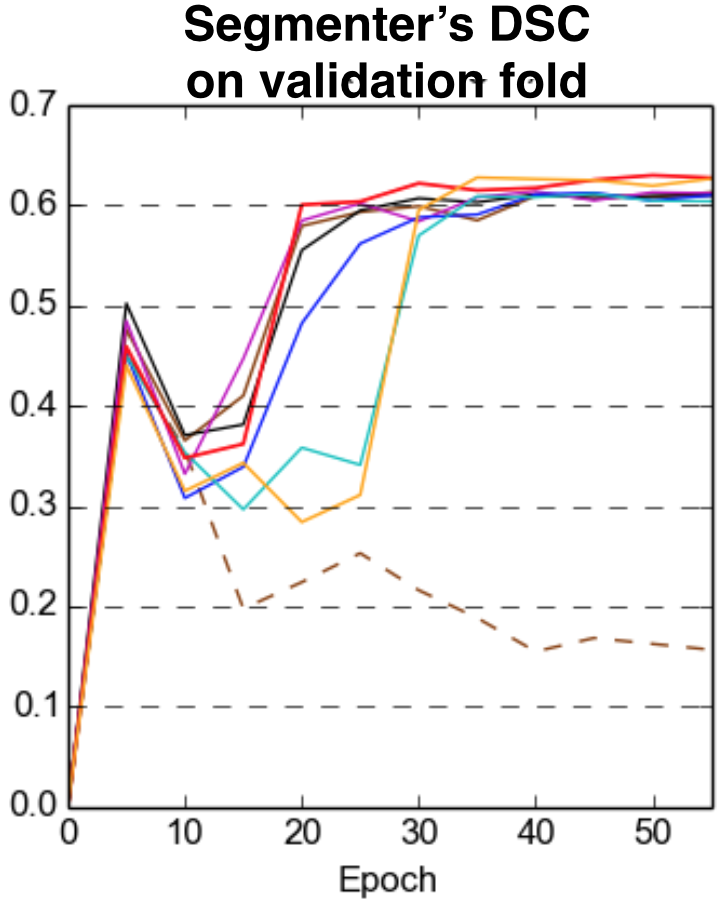}
	%\caption{}
	%\label{fig:connsC}
\end{subfigure}
\\[1ex] %Break line. So that next figures go in their own line.
\centering
\begin{subfigure}[b]{1.0\textwidth}
\centering
	\includegraphics[clip=true, trim=0pt 0pt 0pt 0pt, width=0.9\textwidth]{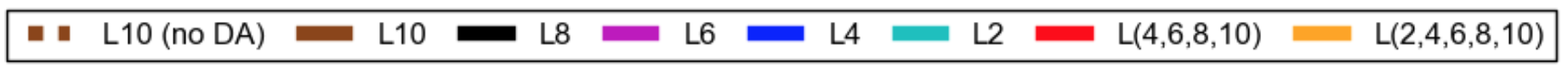}
\end{subfigure}
\caption{Behaviour when the domain-discriminator is connected at different layers of the segmenter. Adaptation is performed after epoch 10 by linearly increasing $\alpha$. Connections at earlier layers lead to higher performance of the discriminator but slower adaptation. Multiple connections increase performance. Note, features learnt at early layers during the refinement in the last stages of training seem more domain-discriminative.}
\label{fig:connections}
\captionof{table}{Final accuracy on $T$ when the discriminator is connected at different depths of the segmenter. Shallow connections increase recall but significantly decrease precision. Multiple connections remove better the source-specific nuisances throughout the segmenter, closing the gap to the practical upper bound of 66.5\% for UDA (Sec.~\ref{subsec:eval}) by approximately 1.5\% DSC. Proposed in bold.}
\scriptsize
\begin{tabular}{@{}llllllll@{}}
\toprule
\multicolumn{1}{c}{}	& L10			& L8				& L6				& L4				& L2				& \textbf{L(4,6,8,10)}	& L(2,4,6,8,10) 		\\ \midrule
DSC					& 61.3(21.0)\hspace{2mm} 	& 61.0(20.7)\hspace{2mm}		& 61.2(19.2)\hspace{2mm}		& 61.0(20.1)\hspace{2mm}		& 60.4(20.2)\hspace{2mm}		& 62.7(19.8)\hspace{2mm}		& 62.7(19.5)		\\
Recall				& 56.9(22.0) 	& 57.3(21.6)		& 57.1(19.8)		& 59.1(20.0)		& 61.1(20.5)		& 58.9(21.2)		& 60.1(20.3)		\\
Precision			& 71.9(20.8) 	& 70.2(20.9)		& 69.9(20.8)		& 68.1(21.6)		& 64.3(21.9)		& 71.6(18.4)		& 69.8(20.0)		\\
\bottomrule
\label{table:connections}
\vspace{-5mm}
\end{tabular}
\end{figure}

%% file: figures/0_figuresToUse/2_fig_effectA/figEffectAInnerLegend.tex
\begin{figure}[t] 
\centering
\label{fig:effectA}
\begin{minipage}[b]{0.45\textwidth}
\caption{The segmenter counters the domain-discriminator after epoch 10, when we linearly increase $\alpha$ from zero to $\alpha_{max}$ until epoch 35. Final accuracy on $T$ was found rather stable for a wide range of values. Decrease greater than 1\% DSC from the highest was found for values 0.02 and 2.0.}
\label{fig:effectA}
\end{minipage}
\begin{subfigure}[b]{0.25\textwidth}
	\centering
	\includegraphics[clip=true, trim=0pt 0pt 0pt 0pt, width=1.0\textwidth]{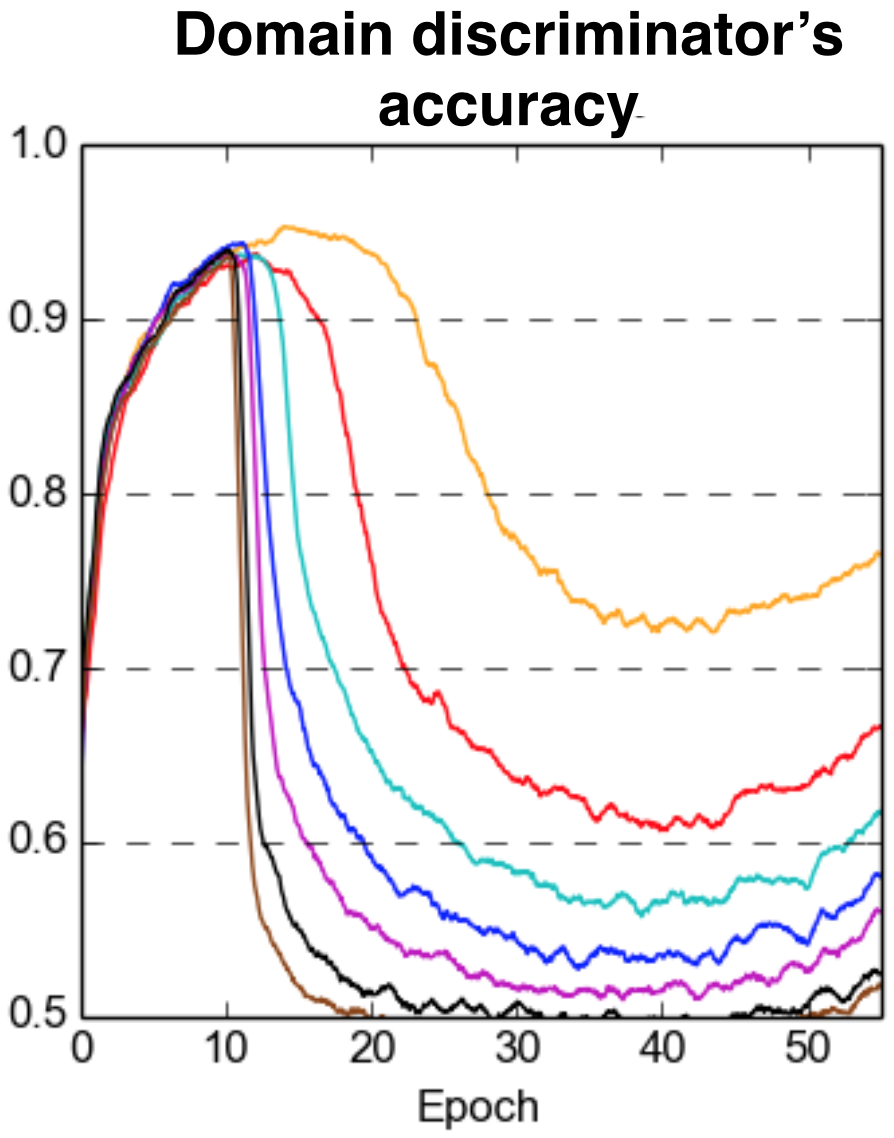}
	%\caption{}
	%\label{fig:connsB}
\end{subfigure}
\begin{subfigure}[b]{0.25\textwidth}
	\centering
	\includegraphics[clip=true, trim=0pt 0pt 0pt 0pt, width=1.0\textwidth]{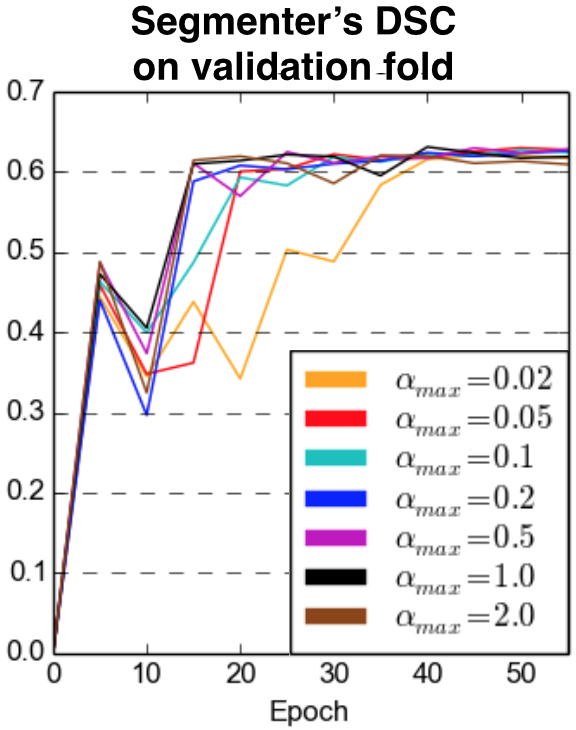}
	%\caption{}
	%\label{fig:connsC}
\end{subfigure}
\vspace{-3mm}
\end{figure}

%% file: sections/conclusion.tex
% !TeX root=../paper.tex

\vspace{-1mm}
\section{Conclusion}
\label{sec:conclusion}
\vspace{-1mm}

We present an unsupervised domain adaptation method for image segmentation by using adversarial training of two 3D neural networks. To the best of our knowledge this is the first work of such an approach on a biomedical imaging problem. Additionally, we propose multi-connected adversarial networks, which perform better by enabling flow of higher quality adversarial gradients throughout the adapted network. We investigate aspects of adversarial training such as the depth of the adapted layer and the strength of adaptation, providing valuable insights for development of future approaches. While unsupervised in the target domain, our method performs close to the accuracy of supervised baselines. We believe our work makes an important contribution in the context of multi-center studies where domain differences are a major limitation in current image analysis methods. Future work will investigate the capabilities of such methods on databases with different types of variations. We also intend to explore domain adaptation via minimization of maximum mean descripancy \cite{borgwardt2006integrating}, which has recently shown competitive results outside the biomedical domain \cite{tzeng2014deep,icml2015_long15}. An implementation of the proposed system will be made publicly available on \url{https://biomedia.doc.ic.ac.uk/software/deepmedic/}.

%% file: sections/acknowledgements.tex
% !TeX root=../journalDeepMedic.tex

%%%%%%%%%%%%%%%%%%%%%%%%%%%%%%%%%%%%%%%%%%%%%%%%%%%%%%%%%%%%%%
%%%%%%%%%%%%%%%%%%%% ACKNOWLEDGEMENTS %%%%%%%%%%%%%%%%%%%%%%%%
%%%%%%%%%%%%%%%%%%%%%%%%%%%%%%%%%%%%%%%%%%%%%%%%%%%%%%%%%%%%%%

\section*{Acknowledgements}

This work is supported by the EPSRC (grant No: EP/N023668/1) and partially funded  by an European Union Framework Program 7 grant (CENTER-TBI; Agreement No: 60215).
Part of this work was carried on when KK was an intern at Microsoft Research Cambridge. KK is also supported by the President's PhD Scholarship of Imperial College London. VN is supported by an Academy of Medical Sciences/Health Foundation Clinician Scientist Fellowship. DM is supported by the Neuroscience Theme of the NIHR Cambridge Biomedical Research Centre and NIHR Senior Investigator awards. We gratefully acknowledge the support of NVIDIA Corporation with the donation of two Titan X GPUs.

%% file: paper.bbl
\begin{thebibliography}{10}
\providecommand{\url}[1]{\texttt{#1}}
\providecommand{\urlprefix}{URL }

\bibitem{ben2010theory}
Ben-David, S., Blitzer, J., Crammer, K., Kulesza, A., Pereira, F., Vaughan,
  J.W.: A theory of learning from different domains. Mach. learning  79(1-2),
  151--175 (2010)

\bibitem{bermudez2016scalable}
Berm{\'u}dez-Chac{\'o}n, R., Becker, C., Salzmann, M., Fua, P.: Scalable
  unsupervised domain adaptation for electron microscopy. In: MICCAI (2016)

\bibitem{borgwardt2006integrating}
Borgwardt, K.M., Gretton, A., Rasch, M.J., Kriegel, H.P., Sch{\"o}lkopf, B.,
  Smola, A.J.: Integrating structured biological data by kernel maximum mean
  discrepancy. Bioinformatics  22(14),  e49--e57 (2006)

\bibitem{ciompi2015automatic}
Ciompi, F., de~Hoop, B., van Riel, S.J., Chung, K., Scholten, E.T., Oudkerk,
  M., de~Jong, P.A., Prokop, M., van Ginneken, B.: Automatic classification of
  pulmonary peri-fissural nodules in computed tomography using an ensemble of
  2d views and a convolutional neural network out-of-the-box. MedIA  26(1),
  195--202 (2015)

\bibitem{ganin2016domain}
Ganin, Y., Ustinova, E., Ajakan, H., Germain, P., Larochelle, H., Laviolette,
  F., Marchand, M., Lempitsky, V.: Domain-adversarial training of neural
  networks. Journal of Machine Learning Research  17(59),  1--35 (2016)

\bibitem{goodfellow2014generative}
Goodfellow, I., Pouget-Abadie, J., Mirza, M., Xu, B., Warde-Farley, D., Ozair,
  S., Courville, A., Bengio, Y.: Generative adversarial nets. In: NIPS (2014)

\bibitem{havaei2016brain}
Havaei, M., Davy, A., Warde-Farley, D., Biard, A., Courville, A., Bengio, Y.,
  Pal, C., Jodoin, P.M., Larochelle, H.: Brain tumor segmentation with deep
  neural networks. MedIA  (2016)

\bibitem{heimann2013learning}
Heimann, T., Mountney, P., John, M., Ionasec, R.: Learning without labeling:
  Domain adaptation for ultrasound transducer localization. In: MICCAI (2013)

\bibitem{jiang2008literature}
Jiang, J.: A literature survey on domain adaptation of statistical classifiers.
  URL: http://sifaka. cs. uiuc. edu/jiang4/domainadaptation/survey  (2008)

\bibitem{kamnitsas2016efficient}
Kamnitsas, K., Ledig, C., Newcombe, V.F., Simpson, J.P., Kane, A.D., Menon,
  D.K., Rueckert, D., Glocker, B.: Efficient multi-scale 3d cnn with fully
  connected crf for accurate brain lesion segmentation. MedIA  36,  61--78
  (2016)

\bibitem{lee2015deeply}
Lee, C.Y., Xie, S., Gallagher, P., Zhang, Z., Tu, Z.: Deeply-supervised nets.
  In: AISTATS. vol.~2, p.~6 (2015)

\bibitem{Long2014}
Long, J., Shelhamer, E., Darrell, T.: Fully convolutional networks for semantic
  segmentation. In: CVPR (2015)

\bibitem{icml2015_long15}
Long, M., Cao, Y., Wang, J., Jordan, M.: Learning transferable features with
  deep adaptation networks. In: ICML (2015)

\bibitem{moeskops2016deep}
Moeskops, P., Wolterink, J.M., van~der Velden, B.H., Gilhuijs, K.G., Leiner,
  T., Viergever, M.A., Isgum, I.: Deep learning for multi-task medical image
  segmentation in multiple modalities. In: MICCAI (2016)

\bibitem{van2015transfer}
van Opbroek, A., Ikram, M.A., Vernooij, M.W., De~Bruijne, M.: Transfer learning
  improves supervised image segmentation across imaging protocols. TMI  34(5),
  1018--1030 (2015)

\bibitem{pan2010survey}
Pan, S.J., Yang, Q.: A survey on transfer learning. IEEE Transactions on
  knowledge and data engineering  22(10),  1345--1359 (2010)

\bibitem{shenton2012review}
Shenton, M., Hamoda, H., Schneiderman, J., Bouix, S., Pasternak, O., Rathi, Y.,
  Vu, M.A., Purohit, M., Helmer, K., Koerte, I., et~al.: A review of magnetic
  resonance imaging and diffusion tensor imaging findings in mild traumatic
  brain injury. Brain imaging and behavior  6(2),  137--192 (2012)

\bibitem{shin2016deep}
Shin, H.C., Roth, H.R., Gao, M., Lu, L., Xu, Z., Nogues, I., Yao, J., Mollura,
  D., Summers, R.M.: Deep convolutional neural networks for computer-aided
  detection: Cnn architectures, dataset characteristics and transfer learning.
  TMI  35(5),  1285--1298 (2016)

\bibitem{tzeng2014deep}
Tzeng, E., Hoffman, J., Zhang, N., Saenko, K., Darrell, T.: Deep domain
  confusion: Maximizing for domain invariance. arXiv preprint arXiv:1412.3474
  (2014)

\bibitem{ullman2016atoms}
Ullman, S., Assif, L., Fetaya, E., Harari, D.: Atoms of recognition in human
  and computer vision. Proc. of the Nat. Academy of Sciences  113(10),
  2744--2749 (2016)

\bibitem{Valiant1984}
Valiant, L.G.: A theory of the learnable. Commun. ACM  27(11),  1134--1142 (Nov
  1984), \url{http://doi.acm.org/10.1145/1968.1972}

\end{thebibliography}
